\documentclass[runningheads]{llncs}
\usepackage{graphicx}
\usepackage{comment}
\usepackage{multirow}
\usepackage{microtype}
\usepackage{csquotes}
\usepackage{appendix}
\usepackage{url}
\begin{document}
\title{Computer Science Named Entity Recognition in \\ the Open Research Knowledge Graph\thanks{Supported by TIB Leibniz Information Centre for Science and Technology, the EU H2020 ERC project ScienceGRaph (GA ID: 819536)}}
%
%
\author{Jennifer D'Souza\orcidID{0000-0002-6616-9509} \and
S\"oren Auer\orcidID{0000-0002-0698-2864}}
\authorrunning{D'Souza and Auer}
%
\institute{TIB Leibniz Information Centre for Science and Technology, Hannover, Germany \\
\email{\{jennifer.dsouza,auer\}@tib.eu}}
\maketitle              
\begin{abstract}
Domain-specific named entity recognition (NER) on Computer Science (CS) scholarly articles is an information extraction task that is arguably more challenging for the various annotation aims that can hamper the task and has been less studied than NER in the general domain. Given that significant progress has been made on NER, we anticipate that scholarly domain-specific NER will receive increasing attention in the years to come. Currently, progress on CS NER -- the focus of this work -- is hampered in part by its recency and the lack of a standardized annotation aim for scientific entities/terms. This work proposes a standardized task by defining a set of seven \textit{contribution-centric scholarly entities} for CS NER viz., \textit{research problem}, \textit{solution}, \textit{resource}, \textit{language}, \textit{tool}, \textit{method}, and \textit{dataset}. 
The main contributions are: 
(1) combining existing CS NER resources that maintain their annotation focus on the set or subset of \textit{contribution-centric scholarly entities} we consider; 
(2) noting the need for big data to train neural NER models, this work additionally supplies thousands of \textit{contribution-centric entity} annotations from article titles and abstracts, thus releasing a cumulative large novel resource for CS NER; and, 
(3) training a sequence labeling CS NER model inspired by state-of-the-art neural architectures from the general domain NER task. 
Throughout the work, several practical considerations are discussed aiming to guide information technology designers of digital libraries.

\end{abstract}
\section{Introduction}

Named entity recognition (NER) is an essential Natural Language Processing (NLP) function for \textit{the lifting of entities-of-interest from unstructured text}. NER powers contemporary knowledge-graph based search engines as demonstrated in industry e.g. by Facebook~\cite{noy2019industry} and Google~\cite{googleKG}; but also the open data community with Wikidata~\cite{vrandevcic2014wikidata}. NER has proven indispensable to machine readers of unstructured texts of common or general knowledge. Commonsense machine reading is an area where significant progress can be tracked via state-of-the-art systems such as Babelfy~\cite{Moroetal:2014iswc}, DBpedia Spotlight~\cite{mendes2011dbpedia}, NELL~\cite{mitchell2018never}, and FRED~\cite{fred}, to name a few. However, the same cannot be said for all domains of discourse text. Consider scholarly literature, as an exemplar, which remains relatively understudied in terms of advanced information retrieval applications that go beyond keywords toward content-based entity-centric machine readers. In the scholarly domain, obtaining fine-grained entity-centric knowledge facilitated by well-established NER systems is not yet feasible. As significant advances have been made on NER in the general domain, we believe that scholarly domain-specific NER will gain increasing attention in the years to come. This is owing to the digitalization of scholarly knowledge impetus via crowdsourcing that is growing~\cite{scigraph,openaire,researchgraph,auer_orkg,baas2020scopus,birkle2020web,wang2020microsoft,hendricks2020crossref}. While expert-based crowdsourcing is effective to obtain high-quality data, it is not necessarily a scalable solution in the face of growing volumes of scientific literature~\cite{sciencegrowthrates}, the processing of which would need the support of automated NLP techniques, one among which is NER. Obtaining the critical digitalized data mass via scalable methods warrant the paradigm shift toward the standardized adoption of digitalized scholarly knowledge. This data representation is advantageous for several reasons, mainly by its meaningful structured connections across entity-centric research progress, research redundancy~\cite{ioannidis2016mass} can be readily alleviated -- a problem predominant in document-based silos of research records where core conceptual entities, buried within text volumes, need to be manually uncovered through human comprehension. Thus, instead of manual human comprehension of the latest and greatest scholarly knowledge within expert silos, digitalized scholarly knowledge can be routinely and centrally screened for information about past and novel discoveries.


Notably, next-generation scholarly digital library (DL) infrastructures are already emerging. The Open Research Knowledge Graph (ORKG)~\cite{orkg} digital research and innovation infrastructure, argues for obtaining a semantically rich, interlinked KG representations of the ``content'' of the scholarly articles, specifically, focused on \textit{research contributions}.\footnote{The ORKG platform can be accessed online: \url{https://orkg.org/}.} With intelligent analytics enabled over such contributions-focused KGs, researchers can track research progress without the cognitive overhead that reading dozens of articles imposes. Via expert crowdsourcing, the information can be readily structured based on human judgements. However, the issue of needing an automated NLP system as a scalable complementary assistance technique remains; one that could even serve the purpose of making it easier for experts to structure scholarly knowledge via drag-and-drop recommendations. A typical dilemma then with leveraging automated NLP for the ORKG, specifically, w.r.t. implementing an NER module is deciding the scholarly/scientific entity extraction targets. In other words, aligned with the ORKG objective of structuring research contributions, the key challenge is: \textit{How to select only the contribution-centric entities and what would be their types}? While this question has a broad scope across Science, in this paper, we tackle the scholarly \textit{contribution-centric} NER problem for Computer Science (CS). 

We define \textit{contribution-centric} CS NER as involving the identification of a word or a phrase as an entity from Computer Science scholarly articles, either from its title, or abstract, or full-text or one or more or all places, which satisfies one of the following seven types, viz. \textit{research problem}, \textit{solution}, \textit{resource}, \textit{language}, \textit{tool}, \textit{method}, and \textit{dataset}. Furthermore, since, in the context of this work, CS NER is qualified as being \textit{contribution-centric}, only those entities that are either the outcome of a particular work or used to support the outcome of that work are candidate extraction targets. Related to the model we propose is the Task, Dataset, Metric (TDM) model attemped in several works \cite{orkg-tdm,scinlp-kg,ibm-tdm} originally inspired from \url{https://paperswithcode.com/}. However, the TDM model targets only the automated construction of Leaderboards in the domain of Artificial Intelligence. In this work, we try to generically define a broader set of \textit{contribution-centric} entities which also naturally subsumes the TDM model.

Summarily, the contributions of this paper are the following. 1) Existing CS NER language resources are examined and the problem of the lack of standardized entities therein is clarified. 2) A standardized set of entities is elicited by proposing a standardization on semantic types for extraction that satisfy the aim of being \textit{contribution-centric} extraction targets for CS. 3) Existing resources that fulfill the aim of \textit{contribution-centric} extraction targets are combined and, further still, additional data is annotated resulting in a large corpus which is made publicly available. 4) Finally, based on empirical evaluations from six different state-of-the-art neural architectures an automated CS NER system is created. The best model performances obtained are between 75\% and 85\% on the task. Our empirical analysis could serve informaticians working in the ecosphere of the contemporary digital libraries. Our system developed for the Open Research Knowledge Graph~\cite{orkg} is also a community release at \url{https://tinyurl.com/orkg-cs-ner}; including its underlying dataset \url{https://github.com/jd-coderepos/contributions-ner-cs}.


\section{Background}

NER is a long-standing task in the NLP community backed by over 3 decades of NLP research tracing it to the Message Understanding Conference series~\cite{muc} where the name was first coined. In contrast, the earliest work on CS NER is relatively recent in 2015 \cite{ftd}. Since MUC~\cite{muc}, NER has seen a broad flurry of activity with numerous scientific events among which are CONLL~\cite{conll02,conll03} and ACE~\cite{ace}. This was also complemented by steady research. NER systems have been variously implemented including handcrafted rules, bootstrapping approaches relying on seed extraction patterns~\cite{collins-singer,etzioni}; as feature-engineered machine learning approaches based on hidden markov models, support vector machines, conditional random fields, and decision trees using features such as orthographic, prefixes suffixes, labels, neighboring words, etc~\cite{hmm,crf-nerc,adaboost,svm,wbi}. Starting with \cite{collobert}, neural network NER systems became popular. The systems differ based on representations relying on words and sub-word units viz. word level architectures~\cite{collobert,drugner}, character level architectures~\cite{kim-char}, character+word level architectures~\cite{chiu-nichols,ma-and-hovy}, and character+word+affix models~\cite{neuroner,yadav}. This mature field is also rife with various tools such as GATE (General Architecture for Text Engineering) \cite{gate}, Mallet (Machine learning for language toolkit) \cite{mccallum}, Natural Language Toolkit (Suite of Python libraries for NLP) \cite{nltk}, DBpedia Spotlight \cite{dbpedia-spotlight}, Stanford NER~\cite{stanfordner}, and SpaCy \cite{spacy}. In contrast, CS NER has only recently started gaining traction following being organized as a task in a SemEval series having then garnered interest in the broader community~\cite{scienceie,gabor-etal-2018-semeval,sciie,ibm-tdm,ai-kg,scirex,ncg}.

\section{Definitions}
\label{sec:definition}

The NER ``named entity'' task from the MUC conferences~\cite{muc} identified the names of all the people, organizations, and geographic locations in a text. Analogously, we define the CS NER task as that of identifying scientific entities of specific semantic types from CS scholarly articles. E.g., the entity ``F1'' of type \textit{metric}; or the entity ``SQuAD'' as an instance of the \textit{dataset} semantic type. In the past~\cite{acl-rd-tec}, the word ``term'' was introduced as a lexical unit carrying a specialised meaning in a particular context. This, we understand, is synonymous to a ``named entity.'' 

Over the years, the CS entity semantic types have evolved w.r.t. the number of types, the type labels, and the extraction target aim they served. Table \ref{comparison-table} shows a high-level overview of the existing semantic types in the context of their datasets. Following which, listed in Table \ref{sem-types} are the nine main semantic types that have emerged, viz. \textit{research problem}, \textit{method}, \textit{solution}, \textit{tool}, \textit{resource}, \textit{dataset}, \textit{language}, \textit{metric}, and \textit{score}. These tables are discussed in detail in Section \ref{cs-ner-rw}. These semantic types are defined as follows~\cite{acl-rd-tec,ibm-tdm,cl-titles}. A \textit{research problem} is the theme of the investigation; a \textit{method} is existing protocols used to support the solution; a \textit{solution} is a novel contribution of a work that solves the \textit{research problem}; entities of type \textit{tool} are found by asking the question ``Using what?'' or ``By which means?;'' \textit{resource} are names of existing data and other references to utilities like the Web, Encyclopedia, etc., used to address the \textit{research problem} or used in the \textit{solution}; \textit{dataset} refers to the name of a dataset; \textit{language} is the natural language focus of a work; a \textit{metric} is the component of evaluation systems used for measuring and measurement processes; and \textit{score} is the system performance number associated with an evaluation \textit{metric}.

\section{Related Work --- Corpora}

This section provides an overview of existing corpora used for scholarly NER research organized by three main research fields.\footnote{Please note access to some datasets may be possible only by contacting the authors.} While this paper only addresses CS NER, in this section the Biomedical and Chemistry domains are undertaken in a related work context to offer a broader perspective on the NER problem.

\subsection{Biomedical NER (BioNER)}

Biomedical NER aims to identify and classify technical terms in the domain of molecular biology for instances of concepts of interest in bioscience and medicine. Examples of such concepts include protein, gene, disease, drug, tissue, body part and location of activity such as cell or organism. Concerning annotated gold-standard corpora to facilitate machine learning, the most frequently used corpora are GENETAG (composed of full text articles annotated with protein/gene mentions) \cite{genetag}, JNLPBA (composed of ~2400 abstracts annotated with DNA, RNA, protein, cell type and cell line concepts) \cite{bionlp}, GENIA (composed of ~200 Medline abstracts annotated with 36 different concepts from the Genia ontology and several levels of linguistic/semantic features) \cite{genia}, NCBI disease corpus (composed of 793 abstracts annotated with diseases in the MeSH taxonomy) \cite{ncbidiseasecorpus}, CRAFT (the second largest corpus consisting of 97 full text papers annotated with over 4000 concepts) \cite{craft} linking to the NCBI Taxonomy, the Protein, Gene, Cell, Sequence ontologies etc. Finally, the MedMentions corpus~\cite{medmentions} posits itself as the largest language resource with over 4000 abstracts annotated with around 34,724 concepts from the UMLS ontology. By leveraging ontologies such as the Gene Ontology \cite{geneontology}, UMLS \cite{umls}, MESH,\footnote{\url{https://www.nlm.nih.gov/mesh/meshhome.html}} or the NCBI Taxonomy \cite{ncbi}, for scientific concept annotations, these corpora build on years of careful knowledge representation work and are semantically consistent with a wide variety of other efforts that exploit these community resources. 

An evolutionary perspective on BioNER shows following endeavors that aimed to establish meaningful semantic structural relations between the bio-molecules such as protein-protein interactions (PPI)~\cite{biocreativeIIppi,biocreativeiiippi}, protein-mutation associations~\cite{mutations}, and gene-disease relations~\cite{genedisease} that were the focus of the BioCreative challenge series; or more complex n-ary bio-molecular events which were addressed in the BioNLP challenge~\cite{bionlp09}. These attempts help curate large-scale knowledge by mining large volumes of articles for PPI databases as MINT \cite{mint} and IntAct \cite{intact} or the more detailed knowledge databases such as pathway~\cite{pathguide} or Gene Ontology Annotation (GOA) \cite{goa}. Further, the ability to construct such connections between bio-molecules also provide a global view on different biological entities and their interactions, such as disease, genes, food, drugs, side effects, pathways, and toxins, opening new routes of research. A more comprehensive survey is available here \url{https://orkg.org/comparison/R163265/}.

\subsection{Chemistry NER (ChemNER)}

BioNER in part fosters Chemistry NER (ChemNER). ChemNER has evolved from extracting drugs~\cite{biocreativeDDI}, to chemicals~\cite{chemdnerCorpus}, chemical disease relations \cite{biocreativeVtaskcorpus}, and drug and chemical-protein interactions \cite{biocreativeVIIchemprot,drugprot}. Text mining for drug and chemical compound entities~\cite{biocreativeDDI,chemdnerCorpus} are indispensable to mining chemical disease relations \cite{biocreativeVtaskcorpus}, and drug and chemical-protein interactions \cite{biocreativeVIIchemprot,drugprot}. Obtaining this structured knowledge has implications in precision medicine, drug discovery as well as basic biomedical research. Corpora for ChemNER are \cite{chemnerdata}'s dataset (42 full-text papers with \texttildelow7000 chemical entities), ChemDNER (10,000 PubMed abstracts with 84,355 chemical entities) \cite{chemdnerCorpus}, and NLM-Chem (150 full-text papers with 38,342 chemical entities normalized to 2,064 MeSH identifiers) \cite{nlmchem}. 

Increasingly, text mining initiatives are seeking out recipes or formulaic semantic patterns to automatically mine machine-actionable information from scholarly articles~\cite{chemrecipes,labprotocols,mysore2019materials,kuniyoshi2020annotating}. In~\cite{labprotocols}, wet lab protocols are annotated, covering a large spectrum of experimental biology, including neurology, epigenetics, metabolomics, cancer and stem cell biology, with actions corresponding to lab procedures and their attributes including materials, instruments and devices. Thereby the protocols then constituted a prespecified machine-readable format as opposed to the ad-hoc documentation norm. Kulkarni et al.~\cite{labprotocols} even release a large human-annotated corpus of semantified wet lab protocols to facilitate machine learning of such shallow semantic parsing over natural language instructions. Within scholarly articles, such instructions are typically published in the Materials and Method section of papers from Biology and Chemistry. \cite{chemrecipes,mysore2019materials} present semantically structured material synthesis procedures capturing synthesis operations (i.e. predicates), and the materials, conditions, apparatus and other entities participating in each synthesis step. These initiatives lend further evidence to the evolution of digital library technologies toward digitalization.

\subsection{Computer Science NER (CS NER)}
\label{cs-ner-rw}

Table \ref{comparison-table} shows existing CS NER corpora compared along five dimensions: (1) domain --- the CS subarea(s) covered in the annotated data, (2) annotation coverage --- the aspect of the paper annotated, (3) scientific entity semantic types --- the semantic types assigned to the entities, (4) size --- the number of papers, tokens, entities in the respective corpora, and (5) annotation method --- the method by which the annotations were obtained. To offer a few summary observations. Most of the corpora consist of relatively short documents. The shortest is the CL-Titles corpus~\cite{cl-titles} with only paper titles. The longer ones have sentences from full-text articles, viz. ScienceIE~\cite{scienceie}, NLP-TDMS~\cite{ibm-tdm}, SciREX~\cite{scirex}, and ORKG-TDM~\cite{orkg-tdm}. We see that the corpora have had from one (e.g., the NCG corpus~\cite{ncg}) to atmost seven entity types (e.g., ACL-RD-TEC~\cite{acl-rd-tec}). Each corpora' types purposefully informs an overarching knowledge extraction objective. E.g., the \textit{focus}, \textit{technique}, and \textit{domain} entity types in the FTD corpus~\cite{ftd} helped examine the influence between research communities; ACL-RD-TEC~\cite{acl-rd-tec} made possible a broader trends analysis with seven types. Eventually, corpora began to shed light on a novel scientific community research direction toward representing the entities as knowledge graphs~\cite{auer_orkg} with hierarchical relation annotations such as synonymy~\cite{scienceie} or semantic relations such `\textit{Method} \textit{Used-for} a \textit{Task}'~\cite{sciie}; otherwise, scientific types were combined within full-fledged semantic constructs as \textsc{Leaderboards} with between three to four concepts~\cite{ibm-tdm,scirex,smallnlp,orkg-tdm}, viz. \textit{research problem}, \textit{dataset}, \textit{method}, \textit{metric}, and \textit{score}; or were in extraction objectives with solely \textit{contribution-centric} entities of a paper~\cite{used_meth_dataset,cl-titles}. Overall, the corpora served two main information extraction aims. 1) Some resources offered all mentions of CS relevant scientific entities: ACL-RD-TEC \cite{acl-rd-tec}, ScienceIE \cite{scienceie}, and STEM-ECR \cite{stem-ecr}; and 2) While others offered only \textit{contribution-centric} entities: FTD \cite{ftd}, SciERC \cite{sciie}, NLP-TDMS \cite{ibm-tdm}, SciREX \cite{scirex}, NCG \cite{ncg}, ORKG-TDM \cite{orkg-tdm}, and CL-Titles \cite{cl-titles}. 

\begin{table}[!htb]
\centering
\caption{Comparison of Computer Science papers corpora for named entity recognition (CS NER). The corpora names in bold are the corpora merged as part of the dataset of this work. Domain Acronyms. CL - Computational Linguistics; CS - Computer Science; MS - Material Science; Phy - Physics; AI - Artificial Intelligence; STEM - Science, Technology, Engineering, Medicine; ML - Machine Learning; CV - Computer Vision. \textbf{Size} column name acronyms: $P$ - papers; $T$ - tokens; $E$ - entities. A detailed version of this table is available here https://orkg.org/comparison/R150058.}
\label{comparison-table}
\begin{tabular}{p{1.5cm}|p{1.2cm}|p{1.5cm}|p{4.4cm}|p{1.2cm}p{1.4cm}p{1cm}} \hline
\multirow{2}{*}{\textbf{Corpora}} & \multirow{2}{*}{\textbf{Domain}} & \multirow{2}{*}{\textbf{Coverage}}    & \multirow{2}{*}{\textbf{Entity Semantic Types}} & \multicolumn{3}{c}{\textbf{Size}} \\ 
 &   &   &   & \textbf{$P$} & \textbf{$T$} & \textbf{$E$}  \\ \hline
\textbf{FTD}~\cite{ftd} & CL & titles, abstracts & focus, domain, technique & 426    & 57,182    & 5,382 \\ \hline
ACL-RD TEC~\cite{acl-rd-tec} & CL & abstracts & language resource, language resource product, measures and measurements, models, other, technology and method, tool and library & 300 & 32,758    & 4,391 \\ \hline
ScienceIE \cite{scienceie} & CS, MS, Phy & full text & material, process, task & 500    & 83,753    & 10,994 \\ \hline
\textbf{SciERC} \cite{sciie} & AI & abstracts & evaluation metric, generic, material, method, task & 500    & 60,749    & 8,089 \\ \hline
NLP-TDMS \cite{ibm-tdm} & CL & titles, abstracts, full text & task, dataset, metric, score & 332 & 1,115,987 & 1,384 \\ \hline
STEM-ECR \cite{stem-ecr} & 10 STEM & abstracts & data, material, method, process & 110    & 26,269    & 6,165 \\ \hline
SciREX \cite{scirex} & ML & titles, abstracts, full text & dataset, method, metric, task & 438    & 248,7091  & \textbf{156,931} \\ \hline
\textbf{NCG} \cite{ncg} & CL, CV & titles, abstracts & research problem & 405 & 47,127 & 908 \\ \hline
ORKG-TDM \cite{orkg-tdm} & AI & titles, abstracts, full text & task, dataset, metric & 5,361 & - & 18,219 \\ \hline
CL-Titles \cite{cl-titles} & CL & titles & language, method, research problem, resource, solution, tool & \textbf{50,237} & 284,672   & 87,567 \\ \hline
\textbf{PwC} (this paper) & AI & titles, abstracts & research problem, method & 12,271 & \textbf{1,317,256} & 29,273 \\ \hline
\textbf{ACL} (this paper) & CL & titles & language, method, research problem, resource, dataset, solution, tool & 31,044 & 263,143   & 67,270 \\ \hline
\end{tabular}
\end{table}

\begin{table}[!htb]
\centering
\caption{Mappings of nine scientific semantic types across Computer Science papers for CS NER. \footnotesize{The first seven italicized types are in the dataset of this work.}}
\label{sem-types}
\begin{tabular}{p{0.3cm}p{3cm}|p{8cm}} \hline
& \textbf{Types}  & \textbf{Mappings in Related Work}  \\ \hline
1 & \textit{research-problem} & domain; application; task; research problem \\ \hline
2 & \textit{method} & technique; technology and method; method \\ \hline
3 & \textit{solution} & focus; solution \\ \hline
4 & \textit{tool} & tool and library; tool \\ \hline
5 & \textit{resource} & language resource; resource \\ \hline
6 & \textit{dataset} & language resource product; dataset \\ \hline
7 & \textit{language} & language \\ \hline
8 & metric & measures and measurements; evaluation metric; metric \\ \hline
9 & score & measures and measurements \\ \hline
\end{tabular}
\end{table}

\section{Our Corpus}

To build our corpus for \textit{contribution-centric} information extraction targets of scientific entities, we aimed: 1) to reuse existing resources for their entity annotations that already fulfill our extraction target aim (described further in Section \ref{existing-res}); and 2) to append additional annotations to create a larger corpus for neural machine learning system development (described in Section \ref{our-resource}).

\subsection{Combining Existing Resources}
\label{existing-res}

This step first entailed normalizing different semantic label names with the same semantic definitions as one standard name. The mappings we used are elicited in Table \ref{sem-types}. The table lists nine main semantic types whose semantic interpretations or definitions were offered in Section 3. After obtaining the label names mappings as semantic type normalizations, we selected only those corpora, and specifically the semantic types within the corpora, that satisfied our CS NER \textit{contribution-centric} entities aim. Overall, our corpus was organized as aggregations of similar parts of the scholarly article. Thus, article titles constitute one corpus called the \textbf{\textsc{Title}s} corpus and article abstracts constitute a second corpus called the \textbf{\textsc{Abstract}s} corpus. Next we describe how some of the existing resources could be reused and combined to form the two respective corpora. 

The \textbf{\textsc{Title}s} corpus combines annotations from two different corpora: 1) the FTD corpus \cite{ftd} (row 1 in Table \ref{comparison-table}) for all three of its entities, viz. \textit{research problem}, \textit{method}, and \textit{solution} entities. In all, 462 titles could be obtained from the FTD corpus which originally also includes exactly 462 total annotated paper titles and abstracts with one or more of the three entities' annotations. And, 2) NCG \cite{ncg} (row 8 in Table \ref{comparison-table}) for its \textit{research problem} entities. In all, 398 titles were obtained from NCG with \textit{research problem} annotations which had a total of approximately 450 papers~\cite{ncg}. Thus the data from these two corpora were merged as the \textbf{\textsc{Title}s} corpus finally containing three entities, viz. \textit{research problem}, \textit{method}, and \textit{solution} deemed by their original corpora respective annotation aims as \textit{contribution-centric}, in turn fulfilling our CS NER aim.

The second corpus, i.e. the \textbf{\textsc{Abstract}s} corpus combines: 1) the FTD corpus paper abstracts annotated with \textit{research problem} and \textit{method} entities. Since no annotations for \textit{solution} entities could be obtained in the abstracts, this type could not be included. As such, abstracts in all 462 of the FTD annotated papers were included in our corpus. Next, 2) 272 abstracts from the NCG corpus with annotations for \textit{research problem}. And, lastly, 3) the SciERC corpus \cite{sciie} (row 4 in Table \ref{comparison-table}) annotated abstracts for its \textit{contribution-centric} \textit{research problem} entity. 431 of its 500 total annotated papers could be combined. Note that SciERC had annotations for additional semantic types as well, e.g., \textit{generic}, \textit{material}, and \textit{method}. However, these annotations could not be included since they did not satisfy our \textit{contribution-centric} entities inclusion criteria. 

\subsection{Our Annotated Data}
\label{our-resource}

\begin{table}[!tb]
\centering
\caption{Inter-annotator percentage agreement scores on 50 titles in the ACL corpus over six scientific semantic entity types}
\label{iaa}
\begin{tabular}{llll} \hline
\textbf{entity type}    & $P$     & $R$     & $F1$     \\ \hline
\textit{solution}         & 86.49 & 71.11 & 78.05 \\
\textit{tool}             & 25    & 16.67 & 20    \\
\textit{dataset}          & 100   & 50    & 66.67 \\
\textit{language}         & 100   & 100   & 100   \\
\textit{method}           & 52.17 & 85.71 & 64.86 \\
\textit{research problem} & 62.96 & 77.27 & 69.39 \\ \hline
TOTAL            &       &       & 69 \\ \hline
\end{tabular}
\end{table}

\subsubsection{ACL} 
\label{acl}

This corpus of Computational Linguistics paper titles was originally released as part of the CL-Titles parser software resources~\cite{cl-titles} and was automatically annotated using the rule-based CL-Titles parser. It included all the titles in the ACL Anthology at a specific download dump timestamp (\url{https://aclanthology.org/anthology.bib.gz}). This corpus was re-reviewed in this work for annotation quality and additional scientific semantic types that should be included. As such we noted that the semantic type \textit{dataset} relevant particularly in the domain of Computational Linguistics was not originally included in the annotated types. We heuristically modified the annotations to include the \textit{dataset} semantic type additionally and we further manually verified as many of the annotations as were possible in a fixed time-frame of 2 weeks. In this time, 31,041 of its 49,728 titles could be verified and amended for incorrect annotations. Thus the new verified and adapted version (which we call simply ACL) includes seven \textit{contribution-centric} entities, viz. \textit{language}, \textit{method}, \textit{research problem}, \textit{resource}, \textit{dataset}, \textit{solution}, and \textit{tool}. See last row in Table 1 for details.

While the corpus verification exercise was done by a single primary annotator (a NLP Postdoc), an IAA exercise for 50 randomly selected titles involving the primary annotator and a secondary ``outsider'' annotator (a NLP PhD candidate) was also conducted to gauge the replicability of the primary annotator's judgements. For this, a relatively straightforward process was followed. The primary annotator created definitions for the considered types with example titles. These were then shown to the ``outsider'' annotator and were available to him as reference material during the annotation task itself. He then proceeded with annotating a randomly selected set of 50 titles which were already annotated by the primary annotator but with the annotations unavailable to him. Thus his annotations of the 50 titles were performed in a blind protocol. Following the completion of the annotation task, inter-annotator agreement (IAA) scores were computed using the Cohen's $\kappa$~\cite{cohen1960coefficient} as well as standard F1 metrics. The results were promising. In terms of Cohen's $\kappa$, they had a strong IAA of 71.52\%. Their IAA scores in terms of the F1 metric are shown in Table~\ref{iaa} with detailed agreement scores breakdown per semantic type. From the Table, we see that the annotators had perfect agreement over identifying entities of type \textit{language}, with least agreement when identifying entities of type \textit{tool}. This latter low agreement score could be ascribed to the second annotator preference of \textit{method} vs. \textit{tool}. 

The 31,041 verified titles were appended to our \textbf{\textsc{Title}s} corpus.

\subsubsection{PwC} A second data source was leveraged from which additional annotations were appended to the \textbf{\textsc{Title}s} corpus and \textbf{\textsc{Abstract}s} corpus, respectively. The data was originally sourced from PapersWithCode (\url{https://paperswithcode.com/}) hence it is referred as \textbf{PwC}. Note that three of the datasets listed in Table \ref{comparison-table}, viz. NLP-TDMS~\cite{ibm-tdm}, SciREX~\cite{scirex}, and ORKG-TDMS~\cite{orkg-tdm}, were indeed subsets of the PwC source. While their papers' subsets selected may have respectively varied, they all unanimously relied on annotation by distance labeling. PwC, itself, is a publicly available leaderboard of tasks in AI predominantly representing the AI NLP and Computer Vision research fields among others such as Knowledge Graph Embeddings, Robotics, etc. They release a public download dump of crowdsourced leaderboards in scholarly articles on research problems in AI annotated w.r.t. \textit{task}, \textit{dataset}, \textit{metric}, \textit{score}, and \textit{method} entities. We downloaded the dump from the online source https://paperswithcode.com/about (timestamp 19-10-2021) and obtained data annotations via distance labeling of their crowdsourced annotations for only the \textit{research problem} and \textit{method} entities (see Table \ref{sem-types} for label mappings of \textit{task} as \textit{research problem}). Note that among their five available entities, three entities, viz. \textit{dataset}, \textit{metric}, \textit{score}, were not considered since they were often not direct mentions in the text but were inferable candidates and hence did not satisfy the sequence labeling objective of this work.

Both our \textbf{\textsc{Title}s} and \textbf{\textsc{Abstract}s} corpora were appended with available PwC annotations for \textit{research problem} and \textit{method} entities. This was done by following two selection criteria: 1) since PwC provided over 50,000 papers, we wanted to select only a subsample of the data to avoid skewing our overall dataset annotations to just the two PwC entities (i.e., \textit{research problem} and \textit{method}), since owing to the \textit{ACL} data (Table \ref{acl}), \textbf{\textsc{Title}s} has six semantic types in all. And 2) we wanted to select a subsample size representative enough of PwC to capture the different nature of their crowdsourced annotations. Table \ref{titles-abstracts-select} provides an insight of how the selection criteria were implemented specifically in terms of how much data was additionally included. Starting with the \textbf{\textsc{Title}s} corpus, the PwC titles were grouped in three categories: those with both \textit{research problem} and \textit{method} mentions; those with either one. From each group, roughly 2000 titles were added to dataset. Similarly, for the \textbf{\textsc{Abstract}s} corpus, the paper abstracts were grouped in three categories: those with both \textit{research problem} and \textit{method} mentions; those with either one. From each of the three groups, roughly 2000 abstracts on average were added to the overall data.

\begin{table}[!th]
\centering
\caption{Selection criteria and statistics for Titles and Abstracts, respectively, from https://paperswithcode.com/ (PwC) that were appended in our overall corpus}
\label{titles-abstracts-select}
\begin{tabular}{|p{4.2cm}|p{1.5cm}||p{4.2cm}|p{1.5cm}|} \hline
\textbf{Titles criteria}    & \textbf{statistics} & \textbf{Abstracts criteria}    & \textbf{statistics} \\ \hline
1, 2, or 3 tasks and 1, 2, or 3 methods & 1,855  & 1, 2, 3, 4, or 5 tasks and methods  & 1,756 \\
1, 2, 3, or 4 tasks and no method for titles & 2,100 & 1, 2, or 3 tasks and no method & 2,500 \\
no task and 1 or 2 methods for titles   & 2,100 & no task and 1, 2, or 3 methods & 2,500 \\ \hline
\end{tabular}
\end{table}

Our final resulting corpus distributions in terms of the constituent corpora after ``combining existing resources'' and adding ``our annotated data'' was as follows. \textbf{\textsc{Title}s} corpus constituent subcorpora distributions: 31,041 (82\%) ACL/5,885 (15\%) PwC/462 (1\%) FTD/398 (1\%) NCG. The sizes of the FTD and NCG are the original dataset sizes. The considered annotations were for seven \textit{contribution-centric} entities, viz. \textit{solution}, \textit{tool}, \textit{dataset}, \textit{language}, \textit{method}, \textit{resource}, and \textit{research problem}. And, the \textbf{\textsc{Abstract}s} corpus constituent subcorpora distributions were: 6756 (85\%) PwC/462 (5\%) FTD/272 (3\%) NCG/431 (5\%) SciERC. While only PwC was a strategically chosen subset for being representative of the two entities, the other corpora were included by their original sizes and \textit{contribution-centric} entities annotations availability. Our corpus is released at \url{https://github.com/jd-coderepos/contributions-ner-cs}.

\section{Our CS NER Sequence Labeler}

\subsection{Experimental Setup}

\paragraph{Model.} Overall, we experimented with six different sequence labeling neural architectures. The basic building blocks to these six architectural variants, inspired from state-of-the-art neural sequence labelers in the general domain~\cite{chiu-nichols,ma-and-hovy,lample-et-al,yang,peters}, were are follows: 1) a CNN, a LSTM, or a BiLSTM first layer over word representations of the data, 2) with and without a second char CNN layer, and 3) an output layer as a CRF decoder since CRFs outperformed the softmax function in sequence labeling tasks. Thus the following six architectures were experimented with: i) word CNN + CRF, ii) word LSTM + CRF, iii) word BiLSTM + CRF, iv) word CNN + char CNN + CRF, v) word LSTM + char CNN + CRF, and vi) word BiLSTM + char CNN + CRF. As mentioned before, each of these architectural configurations, i.e. leveraging only word representations in the first layer or the character-based CNN as the second layer are deconstructions of state-of-the-art sequence labelers for NER in the general domain~\cite{chiu-nichols,ma-and-hovy,lample-et-al,yang,peters}. Further, the word representations for the first layer were computed one of two ways: either directly from the data, or as precomputed vectorized embedding representations.\footnote{We used GloVe embeddings~\cite{glove}.}

For implementing the sequence labelers, we leveraged the open-source toolkit called NCRF++~\cite{ncrf++} (\url{https://github.com/jiesutd/NCRFpp}) that is based on PyTorch. Our experimental configuration files for model hyperparameter details including learning rate, dropout rate, number of layers, hidden size etc., are on Github \url{https://github.com/jd-coderepos/contributions-ner-cs}.

\paragraph{Evaluaton Metrics} We leverage the micro Precision ($P$), Recall ($R$), and F1-score ($F1$) measures for overall task performance. By micro-measures, the total true positive (tp), false positive (fp), true negative (tn), and false negative (fn) counts are computed over all types across all test data instances, i.e. across all titles or across all abstracts, respectively. Following which the precision, recall, and F1 measures are obtained. This helps evaluate task performance at a fine-grained level per semantic type which would take into account low performances on minority semantic types as well. Further, we also leverage the standard $P$, $R$, and $F1$ scores per entity type.

\begin{table}[!tbh]
\centering
\caption{Results with different neural architectures for CS NER over seven semantic concepts with embeddings computed on the data source (top row) and with pretained embeddings (bottom row) on the \textsc{Title}s corpus (columns 2 to 4) and \textsc{Abstract}s corpus (columns 5 to 7).}
\label{cs-ner-titles-abstracts}
\begin{tabular}{p{3.6cm}|p{0.9cm}p{0.9cm}p{0.9cm}|p{0.9cm}p{0.9cm}p{0.9cm}} 
\textbf{Neural Architectures} & \textbf{micro $P$} & \textbf{micro $R$} & \textbf{micro $F1$} & \textbf{micro $P$} & \textbf{micro $R$} & \textbf{micro $F1$} \\ \hline
\small{word CNN + CRF}        & 70.28     & 71.24  & 70.76 & 90.55   & 72.51  & 80.53  \\ \cline{2-7}
                              & 69.32     & 69.16  & 69.24 & 91.78     & 73.58  & 81.68  \\ \hline
\small{word LSTM + CRF}       & 69.24     & 70.08  & 69.65 & 85.45     & 75.54  & 79.62  \\ \cline{2-7}
                              & 68.41     & 66.76  & 67.58 & 90.02     & 71.82  & 79.9  \\ \hline
\small{word BiLSTM + CRF}     & 71.92     & 73.34  & 72.62 & 88.22     & 76.24  & 81.79  \\ \cline{2-7}
                          & 71.44     & 72.91  & 72.17  & 90.14     & 76.36  & 82.68 \\ \hline
\small{word CNN + char CNN}   & 71.31     & 72.96  & 72.13  & 78.61     & 71.08  & 74.65 \\ \cline{2-7}
\small{+ CRF}           & 72.50     & 71.01  & 71.75  & 88.59     & 66.33  & 75.86 \\ \hline
\small{word LSTM + char CNN}        & 72.01     & 72.4   & 72.21  & 85.48   & 78  & 81.57 \\ \cline{2-7}
\small{+ CRF}        & 71.59     & 69.65  & 70.61  & 87.71  & 76.49  & 81.71 \\ \hline
\small{word BiLSTM + char } & \textbf{74.14} & \textbf{76.26} & \textbf{75.18} & \textbf{84.89} & \textbf{81.9} & \textbf{83.37} \\ \cline{2-7}
\small{CNN + CRF}   & 73.67     & 75.16  & 74.41 & 88.2  & 78.85 & 83.26 \\ \hline
\end{tabular}
\end{table}

\begin{table}[!htb]
\centering
\caption{CS NER percentage scores per scientific entity type}
\label{per-entity-results}
\begin{tabular}{llll} \hline
\textbf{Types}  & $P$   & $R$  & $F1$ \\ \hline 
\textit{method}     & 66.8  & 49.13 & 56.62 \\ 
\textit{tool}       & 72.01 & 66.05 & 68.9 \\
\textit{dataset}   & 72.9  & 68.42 & 70.59 \\ 
\textit{research problem} & 68.24 & 79.68 & 73.52 \\
\textit{resource} & 75.72 & 78.61 & 77.14 \\
\textit{solution}  & 78.51 & 82.61 & 80.51 \\
\textit{language}  & 86.22 & 87.78 & 86.99 \\ \hline
\end{tabular}
\end{table}

\subsection{Results and Analysis}

Table~\ref{cs-ner-titles-abstracts} shows the results from our six neural sequence labeling architectural configurations over our two respective corpora, viz. \textbf{\textsc{Title}s} and \textbf{\textsc{Abstract}s}, respectively, for the task of \textit{contribution-centric} CS NER. Our best performing configuration on both datasets is \textit{word-based BiLSTM + character CNN + CRF}. From the first three columns for \textbf{\textsc{Title}s} results, the highest performance is 75.18\% in micro F1 over its seven entities obtained with word embeddings computed directly on the data source. And from the last three columns for \textbf{\textsc{Abstract}s} corpus results, we see the highest performance is 83.37\% in micro F1 over its two semantic types again using word embeddings computed directly on the data source. Their performances are analyzed in detail next.

\paragraph{CNN versus LSTM in the first layer?} From the results, we observe that word-based BiLSTMs outperform word-based CNNs which in turn outperform word-based LSTMs. Thus word-based BiLSTMs are clearly the best neural model for the first layer for \textit{contributions-centric} CS NER. This observation is aligned with the state-of-the-art NER model configuration in the general domain as well.

\paragraph{Is a char CNN layer preferable in the second layer of the sequence learning neural architecture?} We find that it is. Comparing the results in the last three rows with the first three rows of Table~\ref{cs-ner-titles-abstracts}, shows the models discriminative ability significantly increases. This is more evident for the \textbf{\textsc{Title}s} corpus which had seven semantic types compared to the \textbf{\textsc{Abstract}s} corpus with only two. In the former case, a more robust model would be needed. The added character CNN layer satisfies this need.

\paragraph{Is it beneficial to leverage pre-trained embeddings?} We see that it is more beneficial to compute embeddings directly on the dataset rather than using the pretrained embeddings out-of-the-box. However, present advanced embedding models based on transformers such as BERT~\cite{bert} and its variants~\cite{scibert} also allow finetuning the pretrained embeddings on respective experimental datasets. We relegate this experiment to future work. We hypothesize that such embeddings could be leveraged with effective results in a sequence labeling setting as well. However, considering the case presented in this work, i.e. leveraging word embeddings directly computed on a large enough underlying data source versus using pretrained word embeddings, we empirically verify that the former method is better suited to the task.

\paragraph{Which scientific entity type is easiest versus hardest to classify?} These results are shown in Table \ref{per-entity-results}. Viewing the scores of the seven entity types with the \textbf{\textsc{Title}s} and the \textbf{\textsc{Abstract}s} datasets as one combined resource, we find the five semantic types, viz. \textit{dataset}, \textit{research problem}, \textit{resource}, \textit{solution}, and \textit{language} obtained scores above 70\%. Thus for these types our model proves practically suitable. Of these five types, \textit{language} was the highest performing extraction target. This is also consistent with the IAA scores (see Table \ref{iaa}) between the human annotators who annotated the \textit{language} entity type with perfectly matching annotation consistency. The sequence labeler performed worst on \textit{method} entity type. This score discrepancy can be explained by the confusion between the \textit{tool} entity type and \textit{method} entity type which were shown not easily distinguishable during the IAA experiments as well.

\section{Conclusion and Future Directions}

This work has reported a research direction on unifying prior work on scholarly domain-specific NER, specifically for CS NER. It discussed the reuse of existing resources and the complementary addition of new annotations as a contributing publicly available language resource in the community. Furthermore, drawing on observations of state-of-the-art NER systems in the general domain where the NER task itself has garnered much research interest, six neural sequence labeling architectural variants were empirically tested for CS NER. Consequently new machine learning empirical insights could be supplied as a result of this work regarding the strengths of suitable architectural components. We show that the overall CS NER task of extracting \textit{contribution-centric} entities involving seven semantic types has performances above 75\% demonstrating itself as a reliable predictor of entities in practical, real-world system usage settings. The code base is publicly released \url{https://tinyurl.com/orkg-cs-ner}, as well as service calls via a REST API \url{https://tinyurl.com/csner-rest-api} and as a Python package \url{https://tinyurl.com/cs-ner-pypi}.

Given rapid scholarly publication trends~\cite{jinha2010article} --- one that is only further bolstered with the sharing of PDF preprints ahead (or even instead) of peer-reviewed publications~\cite{chiarelli2019accelerating} --- the research community is faced with a crucial dilemma. \textit{How to stay on-track with the past and the current rapid-evolving research progress?} The manual comprehension of this information is nearly impossible for humans. Furthermore, an enormous amount of digital information is expressed as natural-language (NL) text that is not easily processable by computers. Knowledge Graphs (KG) offer a widely used format for representing information in computer-processable form which are increasingly being supported in next-generation scholarly digital library platforms, e.g., ORKG~\cite{orkg}. Such systems currently rely on crowdsourcing methods to obtain good quality data. NLP is therefore needed for mining knowledge graphs from texts. A central part of the problem is to extract the named entities. To this end, this work has taken on the branch of CS NER with a \textit{contribution-centric} entities information extraction aim. As part of future work, the existing set of CS NER entities will be investigated to increase the types coverage. Further, other domains of the Science will also be explored for the scholarly-domain-specific NER task.
%
%
%
%
\bibliographystyle{splncs04}
\bibliography{mybibliography}
\end{document}